# CAUSALITY LEARNING WITH WASSERSTEIN GENERATIVE ADVERSARIAL NETWORKS

Hristo Petkov[1], Colin Hanley[2] and Feng Dong[1]

[1]Department of Computer and Information Sciences, University of Strathclyde, Glasgow, United Kingdom
hristo.petkov@strath.ac.uk
feng.dong@strath.ac.uk

[2]Department of Management Science, University of Strathclyde, Glasgow, United Kingdom
colin_hanley@hotmail.co.uk

## ABSTRACT

*Conventional methods for causal structure learning from data face significant challenges due to combinatorial search space. Recently, the problem has been formulated into a continuous optimization framework with an acyclicity constraint to learn Directed Acyclic Graphs (DAGs). Such a framework allows the utilization of deep generative models for causal structure learning to better capture the relations between data sample distributions and DAGs. However, so far no study has experimented with the use of Wasserstein distance in the context of causal structure learning. Our model named DAG-WGAN combines the Wasserstein-based adversarial loss with an acyclicity constraint in an auto-encoder architecture. It simultaneously learns causal structures while improving its data generation capability. We compare the performance of DAG-WGAN with other models that do not involve the Wasserstein metric in order to identify its contribution to causal structure learning. Our model performs better with high cardinality data according to our experiments.*

## 1. INTRODUCTION

Causal relationships constitute important scientific knowledge. Bayesian Networks (BN) are graphical models representing casual structures between variables and their conditional dependencies in the form of Directed Acyclic Graphs (DAG). They are widely used for causal inference in many application areas such as medicine, genetics and economics [1-3].

Learning causal structure from data is difficult. One of the major challenges arises from the combinatorial nature of the search space. Increasing the number of variables leads to a super-exponential increase of possible DAGs. This makes the problem computationally intractable.

Over the last few years, several methods have been proposed to address the problem of discovering DAGs from data [4], including the score-based methods and constraint-based methods. Kuipers-et-al [5], Heckerman-et-al [6] and Bouckaert [7] have proposed score-based methods (SBMs). They formulate the causal structure learning problem into the optimization of a score function with acyclicity constraint. SBMs rely on the utilization of general optimization techniques to discover DAG structures. Additional structure assumptions and approximate searches are often needed because the complexity of the search space remains super-exponential.

Pearl [8], Spirtes-et-al [1] and Zhang [9] have proposed constraint-based methods (CBMs) which utilize conditional independence tests to reduce the DAG search space and provide graphs that satisfy a set of conditional independencies. However, CBMs often lead to spurious results. Also they are not very robust over sampling noise. Some works have tried to combine SBMs with CBMs. For example, in the MMHC algorithm [10], Hill Climbing is used as the score function and Min-Max Parents and Children (MMPC) is used to check for relations between variables.

Recently Zheng et al. [11] made a breakthrough by transforming the causal structure learning problem from combinatorial to continuous optimization with an acyclicity constraint, which can be efficiently solved using conventional optimization methods. This technique has been extended to cover nonlinear cases. Yu et al. [12] have proposed a model named DAG-GNN based on a variational auto-encoder architecture. Their method can handle linear, non-linear, continuous and categorical data. Another model named GraN-DAG [13] has also been proposed to learn causality from both linear and non-linear data.

Our work is related to the use of generative adversarial networks (GANs) for causal structure learning. GANs is a powerful approach to generate synthetic data. They have also been experimented with to synthesize tabular data [14-15]. To leverage GANs for causal structure learning, a fundamental question is to identify how much the data distribution metrics contribute to causal structure learning. Recently, Gao et al. [16] developed a GAN-based model (DAG-GAN) that learns causal structures from tabular data based on Maximum Mean Discrepancy (MMD) [17]. In contrast, our work has investigated Wasserstein GAN (WGAN) for learning causal structures from tabular data. The Wasserstein distance metric from optimal transport distance [18] is an established metric that preserves basic metric properties [19-22] which has led to Wasserstein GAN (WGAN) to achieve significant improvement in training stability and convergence by addressing the gradient vanishing problem and partially removing mode collapse [23]. However, to the best of our knowledge, so far no study has been conducted to experiment with the Wasserstein metric for causality learning.

Our proposed new DAG-WGAN model combines WGAN-GP with an auto-encoder. A critic (discriminator) is involved to measure the Wasserstein distance between the real and synthetic data. In essence, the model learns causal structure in a generative process that trains the model to realistically generate synthetic data. With the explicit modelling of learnable causal relations (i.e. DAGs), the model learns how to generate synthetic data by simultaneously optimizing the causal structure and the model parameters via end-to-end training. We compare the performance of DAG-WGAN with other models that do not involve the Wasserstein metric in order to identify the contribution from the Wasserstein metric in causal structure learning.

According to our experiments, the new DAG-WGAN model performs better than other models by a margin in tabular data with wide columns. The model works well with both continuous and discrete data while being capable of producing less noisy and more realistic data samples. It can handle multiple data types, including linear, non-linear, continuous and discrete. We conclude that the involvement of the Wasserstein metric helps causal structure learning in the generative process.

## 2. RELATED WORK

DAGs consist of a set of nodes (variables) and directed edges (connections) between the nodes to indicate direct causal relationships between the variables. When a DAG entails conditional independencies of the variables in a joint distribution, the faithfulness condition allows us to recover the DAG from the joint distribution [24]. We perform DAG learning with data distributions that are exhibited in data samples.

## 2.1. Traditional DAG Learning Approaches

There are three main approaches for learning DAGs from data, including the constraint-based, score-based and hybrid approaches.

### 2.1.1. Constraint-based Approach

Constraint-based search methods limit the graph search space by running local independence tests to reduce the combinatorial complexity of the search space [25]. Examples of the constraint-based algorithms include Causal Inference (CI) [8] and Fast Causal Inference (FCI) [26]; [9]. However, typically these methods only produce a set of candidate causal structures that all satisfy the same conditional independencies.

### 2.1.2. Score-based Approach

Score-based methods associate a score to each DAG in the search space in order to measure how well each graph fits the data. Typical score functions include Bayesian Gaussian equivalent (BGe) [5], Bayesian Discrete equivalent (BDe) [6], Bayesian Information Criterion (BIC) [27], Minimum Description Length (MDL) [7]. Assigning a score to each possible graph is difficult due to the combinatorial nature of the problem. As a result, additional assumptions about the DAGs have to be made - the most commonly used ones are bounded tree-width [28], tree-structure [29] and sampling-based structure learning [30-32].

### 2.1.3. Hybrid Approach

Hybrid methods involve both score-based and constraint-based methods for DAG learning. A subset of candidate graphs is generated by using independence tests and they are scored by a score function. One example of the hybrid methods is named Max-Min-Hill-Climbing [10]. Another example is RELAX [33], which introduces "constraint relaxation" of possibly inaccurate independence constraints of the search space.

## 2.2. DAG Learning with Continuous Optimization

Recently, a new approach named DAG-NOTEARS was proposed by Zheng et al. [11] to transform the causal graph structure learning problem from a combinatorial problem into a continuous optimization problem. This facilitates the use of deep generative models for causal structure learning to leverage relations between data sample distributions and DAGs. However, the original DAG-NOTEARS model can only handle linear and continuous data.

Based on DAG-NOTEARS, new solutions have been developed to cover non-linearity. Yu et al. [12] developed DAG-GNN, which performs causal structure learning by using a Variational Auto-Encoder architecture. GraN-DAG proposed by Lachapelle et al. [13] is another extension from DAG-NOTEARS to handle non-linearity. In GraN-DAG, the calculation of the neural network weights is constrained by the acyclicity constraint between the variables. It can work with both continuous and discrete data. Meanwhile, the latest work in DAG-NoCurl from Yu et al. [34] shows that it is also possible to learn DAGs without using explicit DAG constraints.

Notably, DAG-GAN proposed by Gao et al. [16] is one of the latest works that uses GAN for causal structure learning. The work involves the use of Maximum Mean Discrepancy (MMD) [17] in its loss function. The concept of "Information Maximization" [35-37] is incorporated into the GAN framework. The resulting model handles multiple data types (continuous and discrete). However, their experiments have only covered up to 40 nodes in the graph.

## 3. GENERATIVE MODEL ARCHITECTURES

Our model is based on the combination of Auto-encoder (AE) and Wasserstein Generative Adversarial Network (WGAN).

### 3.1. Auto-encoder architecture

The auto-encoder architecture consists of an encoder which takes input data and produces a latent variable $z$ containing encoded features of the input. The latent variable $z$ is then used by a decoder to reconstructs the input data. This can be represented mathematically as:

$$Enc_\phi(x) \Rightarrow z \qquad Dec_\theta(z) \Rightarrow x', \qquad (1)$$

where both the encoder *Enc* and the decoder *Dec* are functions parameterized by $\varphi$ and $\theta$ respectively.

### 3.2. Wasserstein Generative Adversarial Network

A standard generative adversarial network consists of a pair of generator/discriminator that compete against each other. The generator synthesizes data samples from random noise samples $z$. The discriminator determines whether the generated data is real or fake. WGAN is an important improvement over the original GANs. Unlike GANs (in which the discriminator only tells whether the data samples are real or fake), WGAN calculates the distance between the real and fake data samples with Wasserstein Distance. WGAN-GP (Wasserstein GAN with Gradient Penalty) is an improvement of the WGAN model by adding a gradient penalty to satisfy the Lipschitz constraint. The loss function of WGAN-GP can be represented using the following equations:

$$L_D = \underbrace{\mathop{E}_{\tilde{x} \sim P_g}[D(\tilde{x})] - \mathop{E}_{x \sim P_r}[D(x)]}_{\text{Critic loss}} + \underbrace{\lambda \mathop{E}_{\hat{x} \sim P_{\hat{x}}}[(||\nabla_{\hat{x}} D(\hat{x}) - 1||)^2]}_{\text{gradient penalty}},$$
$$L_G = -\mathop{E}_{\tilde{x} \sim P_g}[D(\tilde{x})], \qquad (2)$$

where the discriminator (critic) $D$ needs to satisfy the 1-Lipschitz constraints, $\mathbb{P}r$ and $\mathbb{P}g$ denote the real and synthetic data distributions, and $\mathbb{P}_{\hat{x}}$ is a distribution between real and generated data samples.

## 4. DAG-WGAN

Our proposed DAG-WGAN model involves causal structure in the model architecture by incorporating an adjacency matrix under an acyclicity constraint - see Figure 1. The model has two main components: (1) an auto-encoder which computes the latent representations of the input data; and (2) a WGAN which synthesizes the data with adversarial loss. The decoder of the auto-encoder is used as the generator for the WGAN to generate synthetic data. The encoder is trained with the reconstruction loss while the decoder is trained according to both the reconstruction and adversarial loss. We cover both continuous and discrete data types. The joint WGANs and auto-encoders architecture is motivated by the success of VAE (variational auto-encoder) GAN to better capture data and feature representation [38].

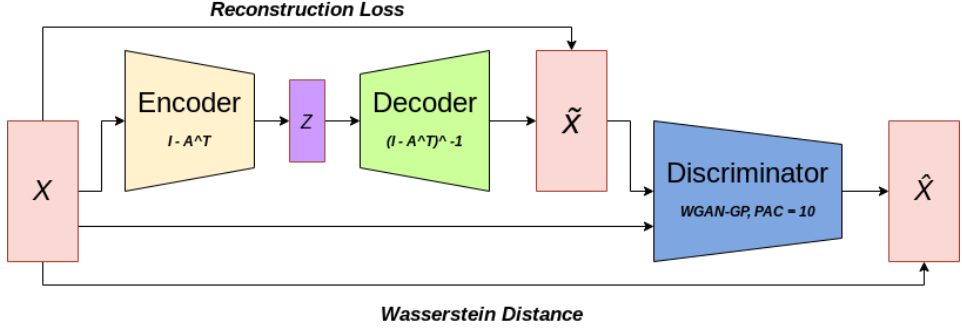

Figure 1. DAG-WGAN Model Architecture

### 4.1. Auto-encoder (AE) and Reconstruction Loss

The use of the auto-encoder is to make sure that the latent space contains meaningful representations, which are used as the noise input to the generator in the training for adversarial loss. The representations are regularized to prevent over-fitting. Similar to Yu et al. [12], we explicitly model the causal structure in both the encoder and decoder by using structural causal model (SCM). The encoder *Enc* is as follows:

$$Enc \equiv Z = (I - A^T)f_1(X) \qquad (3)$$

where $f_1$ is a parameterized function to transform X, $X \in \mathbb{R}^{m \times d}$ is a data sample from a joint distribution of $m$ variables in $d$ dimensions. $Z \in \mathbb{R}^{m \times d}$ is the latent representation. $A \in \mathbb{R}^{m \times m}$ is the weighted adjacency matrix. The corresponding decoder *Dec* is as follows:

$$Dec \equiv X = f_2((I - A^T)^{-1}Z) \qquad (4)$$

where $f_2$ is also a parameterized function that conceptually inverses $f_1$. The functions $f_1$ and $f_2$ can perform both linear and non-linear transformations on Z and X. Each variable corresponds to a node in the weighted adjacency matrix A.

The AE is trained through a reconstruction loss, which can be defined as:

$$L(x, x') = \frac{1}{2} \sum_{i=1}^{m} \sum_{j=1}^{d} (X_{ij} - (M_X)_{ij})^2, \qquad (5)$$

where $M_X$ is a product of the decoder.

To avoid over-fitting, a regularizer is added to the reconstruction loss. The regularizer loss term takes the following form:

$$regularizer = \frac{1}{2} \sum_{i=1}^{m} \sum_{j=1}^{d} (M_Z)_{ij}^2 \qquad (6)$$

where $M_Z$ is the output of the encoder.

### 4.2. WGAN-GP

The decoder of the auto-encoder is also used as the generator of the WGAN model. Alongside the auto-encoder, we use a critic to provide the adversarial loss with gradient penalty. All together, we utilize the critic loss $L_D$ to train the critic, the generator loss $L_G$ (Equation 2) to train the generator (namely the decoder in the AE), and the reconstruction loss $L_R$ (Equation 5) together with the regularizer (Equation 6) to train both the encoder and decoder.

The critic is based upon the popular PacGAN [39] framework with the aim of successfully handling mode collapse and is implemented as follows:

$$\hat{X} = MLP(\tilde{X}, X, leaky - ReLU, Dropout, GP, pac), \quad (7)$$

where $\tilde{x}$ is the data produced from the generator and $X$ is the input data used for the model. *Leaky-ReLU* is the activation function and *Dropout* is used for stability and overfitting prevention. *GP* stands for Gradient Penalty [22] and is used in the loss term of the critic. *Pac* is a notion coming from PacGAN [39] and is used to prevent mode collapse in categorical data, which we found practically useful in terms of improving the outcomes. We use 10 data samples per 'pac' to prevent mode collapse. Note that the number of data samples per 'pac' may vary, however so far we found that 10 produces good results. Furthermore, one might argue that we can remove the 'pac' term entirely when dealing with data which is not categorical. However, our experiment results are not in favour of this idea.

### 4.3. Acyclicity Constraint

Neither minimizing the reconstruction loss, nor optimizing the adversarial loss ensures acyclicity. Therefore, an additional term needs to be added to the loss function of the model. Here we use the acyclicity constraint proposed by Yu et al. [12].

$$tr[(I + \alpha A \circ A)^m] - m = 0, \quad (8)$$

where $A$ is the weighted adjacency matrix of the causal graph, $\alpha$ is a positive hyper-parameter greater, $m$ is the number of the variables, $tr$ is a matrix trace and $\circ$ is the Hadamard product [40] of $A$. This acyclicity requirement associated with DAG structure learning reformulates the nature of the structure learning approach into a constrained continuous optimization. As such, we treat our approach as a constrained optimization problem and use the popular augmented Lagrangian method [41] to solve it.

### 4.4. Discrete Variables

In addition, our model naturally handles discrete variables by reformulating the reconstruction loss term using the Cross-Entropy Loss (CEL) as follows:

$$L(x, x') = -\sum_{i=1}^{m}\sum_{j=1}^{d} X_{ij} log(P_X)_{ij} \quad (9)$$

where $P_X$ is the output of the decoder and $X$ is the input data for the auto-encoder.

# 5. EXPERIMENTS

This section provides experimental results of the model performance by comparing against other related approaches. In particular, our experiments try to identify the contribution from the Wasserstein loss to causal structure learning by making a direct comparison with DAG-GNN [12] where a similar auto-encoder architecture was used without involving the Wasserstein loss. Furthermore, we have also compared the results against DAG-NOTEARS [11] and DAG-NoCurl [34]. All the comparisons are measured using the Structural Hamming Distance (SHD) [42]. More specifically, we measure the SHD between the *learned causal graph* and the *ground truth graph*. Moreover, we also test the integrity of the generated data against CorGAN [14].

The implementation was based on PyTorch [43]. In addition, we used learning rate schedulers and Adam optimizers for both discriminator and auto-encoder with a learning rate of 3-e3.

## 5.1. Continuous data

To evaluate the model with continuous data, our experiments tried to learn causal graphs from synthetic data that were created with known causal graph structures and equations. To allow comparisons, we employed the same underlying graphs and equations like those in the related work, namely DAG-GNN [12], DAG-NoCurl [34] and DAG-NOTEARS [11].

More specifically, the data synthesis was performed in two steps: 1) generating the ground truth causal graph and 2) generating samples from the graph based on the linear SEM of the ground truth graph. In Step (1), we generated an Erdos-Renyi directed acyclic graph with an expected node degree of 3. The DAG was represented in a weighted adjacency matrix $A$. In Step (2), a sample $X$ was generated based on the following equations. We used the linear SEM $X = A^T x + z$ for the linear case, and two different equations, namely $X = A^T h(x) + z$ (non-linear-1) and $X = 2sin(A^T(x+0.5)) + A^T cos(x+0,5) + z$ (non-linear-2) for the nonlinear case.

The experiments were conducted with 5000 samples per graph. The graph sizes used in the experiments were 10, 20, 50 and 100. We measured the SHD (averaged over five different iterations of each model) between the output of a model and the ground truth, and the outcome was compared against those from the related work models (i.e. those mentioned at the beginning of Section 4). In addition to the mean SHD, confidence intervals were also measured based on the variance in the estimated means. These provide insight into the consistency of the model. Tables 1-3 show the results on continuous data samples:

Table 1. Comparisons of DAG Structure Learning Outcomes between DAG-NOTEARS, DAG-NoCurl, DAG-GNN and DAG-WGAN with Linear Data Samples

| Model | SHD (5000 linear samples) | | | |
|---|---|---|---|---|
| | d = 10 | d = 20 | d = 50 | d = 100 |
| DAG-NOTEARS | 8.4 ± 7.94 | 2.6 ± 1.84 | 25.2 ± 19.82 | 106.56 ± 56.51 |
| DAG-NoCurl | 7.9 ± 7.26 | 2.5 ± 1.93 | 24.6 ± 19.43 | 99.18 ± 55.27 |
| DAG-GNN | 6 ± 7.77 | 3.2 ± 1.6 | 21.4 ± 14.15 | 88.8 ± 47.63 |
| **DAG-WGAN** | **2.2 ± 4.4** | **2 ± 1.1** | **4.8 ± 4.26** | **28.20 ± 12.02** |

Table 2. Comparison of DAG Structure Learning Outcomes between DAG-NOTEARS, DAG-NoCurl, DAG-GNN and DAG-WGAN with Non-Linear Data Samples 1

| Model | SHD (5000 non-linear-1 samples) | | | |
|---|---|---|---|---|
| | d = 10 | d = 20 | d = 50 | d = 100 |
| DAG-NOTEARS | 11.2 ± 4.79 | 19.3 ± 3.14 | 53.7 ± 11.39 | 105.47 ± 13.51 |
| DAG-NoCurl | 10.4 ± 4.42 | 17.4 ± 3.27 | 51.6 ± 11.43 | 105.7 ± 13.65 |
| DAG-GNN | **9.40 ± 0.8** | **15 ± 3.58** | 49.8 ± 7.03 | 104.8 ± 12.84 |
| **DAG-WGAN** | 9.8 ± 2.4 | 16 ± 5.4 | **40.40 ± 10.97** | **80.40 ± 9.09** |

Table 3. Comparison of DAG Structure Learning Outcomes between DAG-NOTEARS, DAG-NoCurl, DAG-GNN and DAG-WGAN with Non-Linear Data Samples 2

| Model | SHD (5000 non-linear-2 samples) | | | |
|---|---|---|---|---|
| | d = 10 | d = 20 | d = 50 | d = 100 |
| DAG-NOTEARS | 9.8 ± 2.61 | 22.9 ± 2.14 | 38.3 ± 13.19 | 125.21 ± 61.19 |
| DAG-NoCurl | 7.4 ± 2.78 | 17.6 ± 2.25 | 33.6 ± 12.53 | 116.8 ± 62.3 |
| DAG-GNN | 2.6 ± 2.06 | 3.80 ± 1.94 | 13.8 ± 6.88 | 112.2 ± 59.05 |
| **DAG-WGAN** | **1 ± 1.1** | **3.4 ± 2.06** | **12.20 ± 7.81** | **20.20 ± 11.67** |

## 5.2. Benchmark discrete data

To evaluate the model with discrete data, we used the benchmark datasets available at the Bayesian Network Repository https://www.bnlearn.com/bnrepository/ . The repository provides a variety of datasets together with their ground truth graphs (Discrete Bayesian Networks, Gaussian Bayesian Networks and Conditional Linear Gaussian Bayesian Networks) in different sizes (Small Networks, Medium Networks, Large Networks, Very Large Networks and Massive Networks). To test the scalability of our model, we used datasets of multiple sizes. The datasets utilized in the experiment were Sachs, Alarm, Child, Hailfinder and Pathfinder. The SHD metric was used to measure the performance. Table 4 contains the results from the experiment.

Table 4. Comparison of DAG Structure Learning Outcomes between DAG-WGAN and DAG-GNN with Discrete Data Samples

| Dataset | Nodes | SHD | |
| --- | --- | --- | --- |
| | | DAG-WGAN | DAG-GNN |
| Sachs | 11 | 17 | 25 |
| Child | 20 | 20 | 30 |
| Alarm | 37 | 36 | 55 |
| Hailfinder | 56 | 73 | 71 |
| Pathfinder | 109 | 196 | 218 |

### 5.3. Data integrity

DAG-WGAN was also evaluated by comparing its data generation capabilities against other models. More specifically, we compare the data generation capabilities of the models on a 'dimension-wise probability' basis by measuring how well these models learn the real data distributions per dimension. We used the *MIMIC-III* dataset [44] in the experiments as the same dataset was also used in other comparable works. The data is presented in the form of a patient record, where each record has a fixed size of 1071 entries. We use the same data and pre-processing steps, in order to ensure a fair comparison between all models.

Figure 2 depicts the results of the experiment. We have only compared with CorGAN [14] as it out-performs the other similar models such as medGAN [45] and DBM [46] - see [14] where results of the other models are available. We present the results in a scatter plot, where each point represents one of the 1071 entries and the x and y axes represent the success rate for real and synthetic data respectively. In addition, we use a diagonal line to mark the ideal scenario.

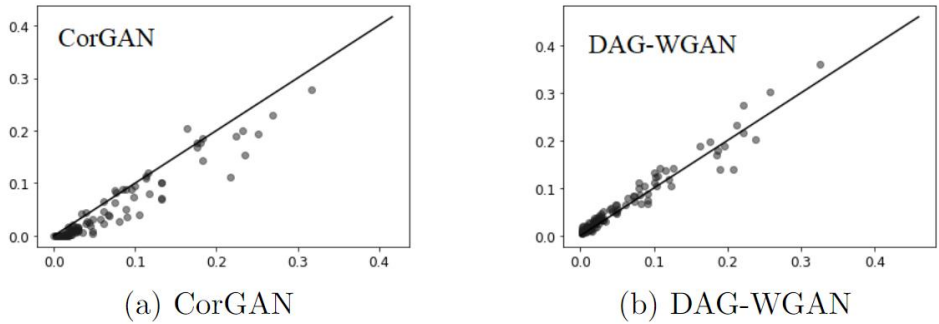

(a) CorGAN     (b) DAG-WGAN

Figure 2. Data generation test results

### 6. DISCUSSION

We discuss the results of both continuous and discrete datasets. The limitations of the model are also addressed together with the plans for future work.

The results on the continuous datasets are competitive across all three cases (linear, non-linear-1 and non-linear-2). According to Tables 1, 2 and 3, our model dominates DAG-NOTEARS and DAG-NoCurl, producing better results throughout all experiments and outperforming DAG-GNN in most of the cases.

For small-scale datasets (e.g. d=10 and d=20), our model performs better than DAG-GNN in most cases and is superior to DAG-NOTEARS and DAG-NoCurl in all the experiments.

For large-scale datasets (e.g. d=50 and d=100), our model outperforms all the other models used in the study by a significantly large margin, which implies that the model can scale better, which is a significant advantage.

Notably, our experiments have mainly focused on the comparisons with DAG-GNN, as we aim to identify contributions from the Wasserstein loss to causal structure learning. In DAG-GNN, a very similar auto-encoder architecture was employed and DAG-WGAN has added WGAN as an additional component. Hence the comparison is meaningful in order to identify contributions from the Wasserstein metric.

For the discrete case, the results from the comparison between the two models are competitive. Out of the five experiments conducted during the study, four results were clearly in favour of our model (i.e.DAG-WGAN) and in one case DAG-GNN was slightly better. These are encouraging results compared to the state of the art model of DAG-GNN.

Also, according to the results illustrated in Figure 2, dimension-wise, the data generated using DAG-WGAN is more accurate and of higher quality than the ones generated using CorGAN, medGAN or DBM.

These results show that DAG-WGAN can handle both continuous and discrete data effectively. They have also demonstrated the quality of the generated data. As the improvement was achieved by introducing the Wasserstein loss in addition to the auto-encoder architecture, the comparisons between them show that the hypothesis on the contribution from the Wasserstein metric to causal structure learning stands.

However, the discrepancy which occurred in the synthetic continuous data results provides us with an insight into the limitations of the model. Some of our early analysis shows that further improvement can be achieved by generalizing the current auto-encoder architecture. Furthermore, as it stands currently, our model does not handle vector or mixed-typed data. These aspects will be further experimented with and reported in our future work.

On the topic of potential improvements, the capability of recovering latent representation places the generative models in a good position to address the hidden confounder challenges in causality learning - some earlier work from [39-40] have moved towards this direction. We will further investigate whether DAG-WGAN can contribute. Last but not least, the latest work in DAG-NoCurl [29] shows that the speed performance can be improved by avoiding the DAG constraints. We will investigate how this new development can be adapted to DAG-WGAN to improve its overall performance.

The capability of generating latent representations places the generative models in a good position to address the hidden confounder challenges in causality learning - some earlier work from [47-48] have moved towards this direction. In the future work, we will further investigate whether DAG-WGAN can contribute to this by incorporating the "Information Maximization" concept [35-37] using the Maximum Mean Discrepancy loss term (MMD) [17]. The inclusion of the MMD loss term to the auto-encoder loss would allow for more meaningful disentangled latent representations, resulting in a potential increase in accuracy. Last but not least, the latest work in DAG-NoCurl [34] shows that the speed performance can be improved by avoiding explicit DAG constraints. We will investigate how this new development can be adapted to DAG-WGAN to improve its overall performance.

# 7. CONCLUSION

In this work, we investigate the contributions of the Wasserstein distance metric to causal structure learning from tabular data. We test the hypothesis with evidence that the Wasserstein metric can simultaneously improve structure learning while generating more realistic data samples. This leads to the development of a novel approach to DAG structure learning, which we coined DAG-WGAN. The effectiveness of our model has been demonstrated in a series of tests, which show the inclusion of the Wasserstein metric can indeed improve the outcomes of causal structure learning. According to our results, the Wasserstein metric allows our model to generate less noisy and more realistic output while being easier to train. The increase in quality of the synthesized data in turn leads to an improvement in structure learning.

## Authors

**Hristo Petkov** received the B.Sc. (First Class) degree from the Department of Computer and Information Sciences, University of Strathclyde. He is currently pursuing the Doctor's degree with the same department. His research interests include medicine, healthcare, deep learning, neural network models.

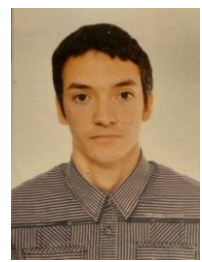

**Colin Hanley** received the M.Sc. degree from the Department of Management Science, University of Strathclyde. Currently, he is in pursuit of a career as a Full Stack Data Developer. His interests include quantitative analysis, human decision making, data analytics.

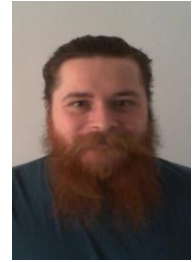

**Feng Dong** is a Professor of Computer Science at the University of Strathclyde. He was awarded a PhD in Zhejiang University, China. His recent research has addressed human centric AI to support knowledge discovery, visual data analytics, image analysis, pattern recognition.

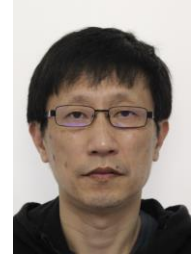